\documentclass[conference]{IEEEtran}
\pdfoutput=1

\IEEEoverridecommandlockouts
\usepackage{cite}
\usepackage{amsmath,amssymb,amsfonts}
\usepackage{algorithmic}
\usepackage{graphicx}
\usepackage{textcomp}
\usepackage{xcolor}
\usepackage{soul}
\usepackage{url}
\usepackage{array}

\def\BibTeX{{\rm B\kern-.05em{\sc i\kern-.025em b}\kern-.08em
    T\kern-.1667em\lower.7ex\hbox{E}\kern-.125emX}}

\newcommand{\hide}[1]{}

\begin{document}

\title{Evaluation of a Skill-based Control Architecture for a Visual Inspection-oriented Aerial Platform\\
\thanks{\textcopyright{} 2020 IEEE. Personal use of this material is permitted. Permission from IEEE must be obtained for all other uses, in any current or future media, including reprinting/republishing this material for advertising or promotional purposes, creating new collective works, for resale or redistribution to servers or lists, or reuse of any copyrighted component of this work in other works.}%
\thanks{This work is partially supported by EU-H2020 projects BUGWRIGHT2 (GA 871260) and ROBINS (GA 779776), and by projects PGC2018-095709-B-C21 (MCIU/AEI/FEDER, UE), and PROCOE/4/2017 (Govern Balear, 50\% P.O. FEDER 2014-2020 Illes Balears). This publication reflects only the authors views and the European Union is not liable for any use that may be made of the information contained therein.}
}


\author{\IEEEauthorblockN{Emilio Garcia-Fidalgo, Francisco Bonnin-Pascual, Joan P. Company-Corcoles and Alberto Ortiz}
\IEEEauthorblockA{\textit{Department of Mathematics and Computer
Science}, \textit{University of the Balearic Islands} and IDISBA, Palma, Spain \\
\{emilio.garcia, xisco.bonnin, joanpep.company, alberto.ortiz\}@uib.es
}
}

\maketitle

\begin{abstract}
The periodic inspection of vessels is a fundamental task to ensure their integrity and avoid maritime accidents. Currently, these inspections represent a high cost for the ship owner, in addition to the danger that this kind of hostile environment entails for the surveyors. In these situations, robotic platforms turn out to be useful not only for safety reasons, but also to reduce vessel downtimes and simplify the inspection procedures. Under this context, in this paper we report on the evaluation of a new control architecture devised to drive an aerial platform during these inspection procedures. The control architecture, based on an extensive use of behaviour-based high-level control, implements visual inspection-oriented functionalities, while releases the operator from the complexities of inspection flights and ensures the integrity of the platform. Apart from the control software, the full system comprises a multi-rotor platform equipped with a suitable set of sensors to permit teleporting the surveyor to the areas that need inspection. The paper provides an extensive set of testing results in different scenarios, under different operational conditions and over real vessels, in order to demonstrate the suitability of the platform for this kind of tasks.
\end{abstract}


\section{Introduction}
\label{sec:intro}

Vessels are one of the most used methods for transporting goods around the world and, therefore, their importance for the international commerce is beyond question. All vessels, irrespectively from their type (e.g. bulk carriers, container-ships, oil-tankers, etc.) and the specific transportation task they implement, are subject to failure because of being affected by different kinds of defects throughout its life. In order to avoid that undetected defects derive into maritime disasters (because of e.g. the vessel structure's being buckled or fractured), they all must be submitted to periodic inspections, sometimes as an initiative of the owner and sometimes due to regulatory requirements (overseen by the so-called Classification Societies). Among the diversity of defective situations which can arise, coating breakdown, corrosion, material thickness diminution and ultimately cracks are typically considered as primary indicators of the state of the hull, and therefore must be properly searched out.

Vessel inspections take place mostly in shipyards (and usually in dry-docks), where proper access means --scaffolding, portable ladders and/or cherry pickers-- must be arranged to permit the surveying personnel to reach all the vessel areas and structures that need scrutinisation, i.e. be within arm's reach from structures. This procedure represents an important expense for the ship owner, due to the fact that the vessel is not under operation and so opportunity costs result. Furthermore, during inspections, surveyors can need to reach high-altitude areas or enter in hazardous environments, jeopardizing their own safety. Because of all this, it is clear the interest of enhancing inspection procedures so that they can be performed as intensively as needed and from a safe position. In this regard, robotic platforms with different locomotion mechanisms can be of application at this point. Of particular relevance are their ability to access places which are usually hard to reach for humans. This is in particular the case of Micro-Aerial Vehicles (MAVs), which, in the last decade, have emerged as a powerful solution within the context of the inspection and monitoring of industrial facilities, mostly motivated by their fast deployment times and reduced size in comparison with other solutions. By way of illustration, MAVs have been shown effective to inspect, among other infrastructures, power plant boilers~\cite{burri2012,nikolic2013}, dam walls and penstocks~\cite{ozaslan2015}, bridges~\cite{jimenez-cano2015}, power lines~\cite{araar2014}, wind turbines~\cite{stokkeland2015}, mines and tunnels~\cite{gohl2014}, petrochemical facilities~\cite{huerzeler2012}, and large-tonnage vessels~\cite{ortiz2016_SENSORS,fang2017}. The reader is referred to~\cite{bonnin-pascual2019_OcEng} for a detailed survey on existing MAVs for inspection.


Despite their proven usefulness for inspection tasks, manoeuvring these platforms by non-skilled users with previous experience, as might be the case of the surveying staff, can easily become a hard task. A MAV designed for inspection tasks would therefore benefit from a control architecture that simplifies its operation, allowing the surveyor to focus on assessing the condition of the vessel structures, while ensuring the platform integrity at the same time. With this in mind, in this paper we describe a novel control architecture, developed in the context of the Supervised Autonomy (SA) paradigm~\cite{cheng2001}, which, by means of intensive use of behaviour-based high-level control, defines a platform fitted with visual-inspection functionalities, several operation modes and different autonomy levels. Furthermore, the brand-new aerial platform \hide{\textit{MUSSOL} (owl, in Catalan)}is equipped with a set of sensors which allows estimating the state of the vehicle and control its motion effectively within the intended operating scenarios, while gathering inspection data during flight. Most importantly, this paper reports on a number of experiments performed within different scenarios and under diverse operational conditions in order to show the suitability of the control architecture for the intended task.


The aerial platform and the control architecture here described have been developed within the context of the EU-funded H2020 ROBINS project\footnote{\url{www.robins-project.eu}}, which, among other goals, intends to fill the technology gap that today still represent a barrier to adopt Robotics and Autonomous Systems (RAS) in activities related to the inspection of ships. The framework proposed results from the experience acquired by our research team as partners in previous projects also related to vessel inspection, such as the EU-funded FP7 MINOAS~\cite{ortiz2010,eich2014} and INCASS~\cite{ortiz2017_JNR} projects. Lessons learnt during these projects, referring to structure, software organization, control approach, platform localization and navigation, and interface with the user, have been taken into account and integrated within the \hide{MUSSOL}system. 


The rest of the paper is organized as follows: Section~\ref{sec:system} overviews the whole system; Section~\ref{sec:mav} details the aerial platform and the sensors it is fitted with; Section~\ref{sec:carch} describes the control architecture, addresing the control approach and the organization of the software; Section~\ref{sec:expres} reports on the results of an extensive set of experiments focusing on the evaluation of the different skills available through the control software; Section~\ref{sec:conclusions}, finally, concludes the paper.

\section{System Overview}
\label{sec:system}


The overall system has been designed around the Supervised Autonomy (SA) paradigm~\cite{cheng2001} in order to provide the platform with mechanisms to ensure its own safety while reducing the effort required from the operator to manoeuvre it. This approach defines a framework for human-robot interaction where the user is not in charge of the complete control of the system, and hence he/she can be focused on the inspection task itself. The SA framework comprises five concepts: \textit{self-preservation}, which states that the robot has to carry out the tasks that may be required to ensure its own integrity; \textit{instructive feedback}, to provide the user with proper perception capabilities, from the data supplied by the robot; \textit{qualitative instructions}, to control the platform in an easy-to-understand way; \textit{qualitative explanations}, to inform the user about what is happening using a qualitative style, at a level compatible with the platform commanding style; and, finally, a \textit{user interface} to visualize the instructive feedback and to allow the user to issue qualitative instructions.

\begin{figure}[tb]
\centering
\begin{tabular}{@{\hspace{0mm}}c@{\hspace{2mm}}c@{\hspace{0mm}}}
\includegraphics[trim={0 0 0 0},clip,width=0.57\linewidth]{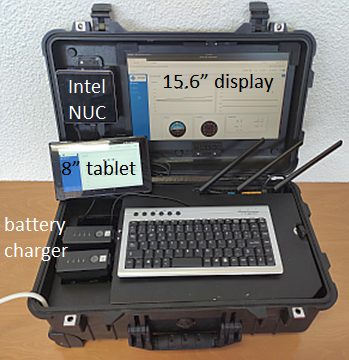} &  
\includegraphics[trim={0 0 35 0},clip,width=0.40\linewidth]{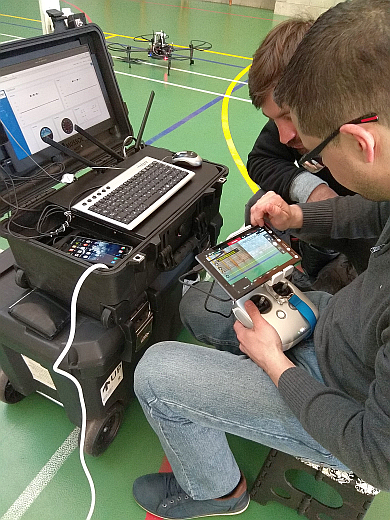} \\
\end{tabular}
\caption{The Base Station to operate the platform. (Left) Ground Control Unit (including an Intel NUC computer, a 15.6 inch. screen, a keyboard and mouse, a dual-band router and three battery chargers) and a tablet PC showing the GUI. (Right) A human operator using the base station and providing commands to the MAV through the remote controller.}
\label{fig:bs}
\end{figure}

In order to implement the SA framework, the system comprises two agents: the MAV and the Base Station (BS). On the one hand, the MAV, equipped with the adequate set of sensors, executes: (1) the control-related tasks required to successfully perform the inspection mission; and, (2) tasks required to ensure its self-preservation. On the other hand, the BS allows the operator to interact with the platform by means of qualitative instructions using a suitable input device, and, at the same time, supplies the available information about the state of the platform and the current mission. This bidirectional communication between the MAV and the BS is performed wirelessly. The BS, shown in Fig.~\ref{fig:bs}, comprises the Ground Control Unit (GCU), one or more remote controllers (R/C) and a web-based Graphical User Interface (GUI).

To finish with this general description of the system, the platform integrates the operator into the main control loop, giving rise to different autonomy levels. This means that, even during the execution of missions that can run in a total autonomous way, the user can take control of the platform whenever necessary to carry out specific tasks, such as, for instance, attaining a difficult-to-reach point of the environment, being assisted at all times by the control software lower control-levels. As an additional benefit of the SA paradigm, the autonomy of the platform can be extended with new functions, as desired, to provide additional, enhanced levels of assistance for infrastructures inspection.

\section{The Aerial Platform}
\label{sec:mav}

Performing an inspection mission requires an accurate estimation of the platform state during the flight. As it is known, GPS signal reception is typically poor or totally unavailable within enclosed spaces, so that platform localization should not rely on it. Due to this reason, the state of the vehicle must be estimated using an alternative and suitable set of sensors, the platform must be fitted with. For obvious reasons, when choosing these sensors, lightweight devices are preferred due to payload limitations.

\begin{figure}[tb]
\centering
\includegraphics[trim={0 13cm 0 13cm},clip,width=1.0\linewidth]{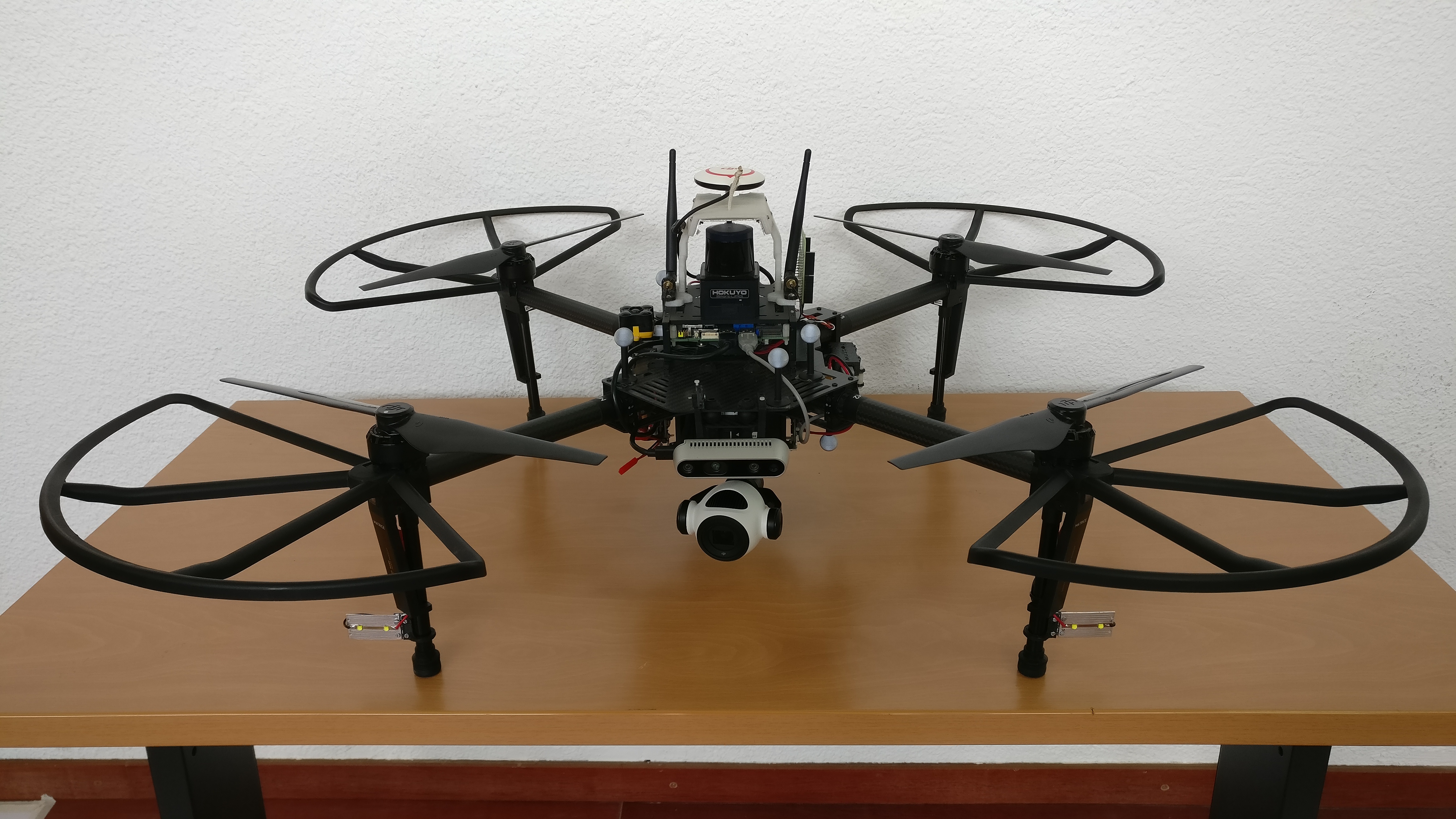}
\caption{Overall view of the aerial platform}
\label{fig:mav}
\end{figure}


The mechanical structure of the vehicle is based on a commercial platform which has been adapted to the purposes of the inspection application. In more detail, the base platform is the Matrice 100 quadrotor manufactured by DJI (see Fig.~\ref{fig:mav}). This vehicle is fitted, by default, with a Flight Management Unit (FMU) for platform stabilization in roll, pitch and yaw, and thrust control. This FMU is equipped with a 3-axis Inertial Measurement Unit (IMU) and a barometric pressure sensor. Additionally, the original configuration supplied by the platform’s manufacturer comprises a GPS receiver and a compass.


As a complement to the default configuration, the platform has been equipped with the following sensors:
\begin{itemize}
    \item A \textit{Hokuyo UST-20LX} laser scanner, used to estimate 2D speed and to measure the distance to surrounding obstacles.
    \item A \textit{LIDAR-Lite v2} laser range finder, pointing downwards, to supply height data. The final platform height is estimated from the fusion of this height with the height provided by the barometric pressure sensor.
    \item A \textit{TeraRanger Evo 60m} infrared range finder, pointing upwards, to supply the distance to the ceiling.
    \item A flexible, dimmable illumination system, consisting of several groups of LEDs pointing in different directions, to properly and adaptively light up the inspection area.
    \item An \textit{Intel Realsense D435i} camera to collect colour and depth images.
    \item A \textit{Zenmuse X3} camera to gather, from the vessel structures under inspection, video footage.
    \item An Ultra-Wide Band (UWB) \textit{Pozyx} receiver/tag, used as part of an UWB-based global localization system (see~\cite{bonnin-pascual2019_ETFA} for additional details).
    \item An embedded computer (featuring an \textit{Intel Core i7-8650U} processor with 16GB RAM). Among other functionalities, this PC executes the critical control loops of the vehicle, hence avoiding to send critical sensor data to the BS and have to account for communications latency.
\end{itemize}

\section{The Control Architecture}
\label{sec:carch}

\begin{figure}[tb]
\centering
\includegraphics[width=1.0\linewidth]{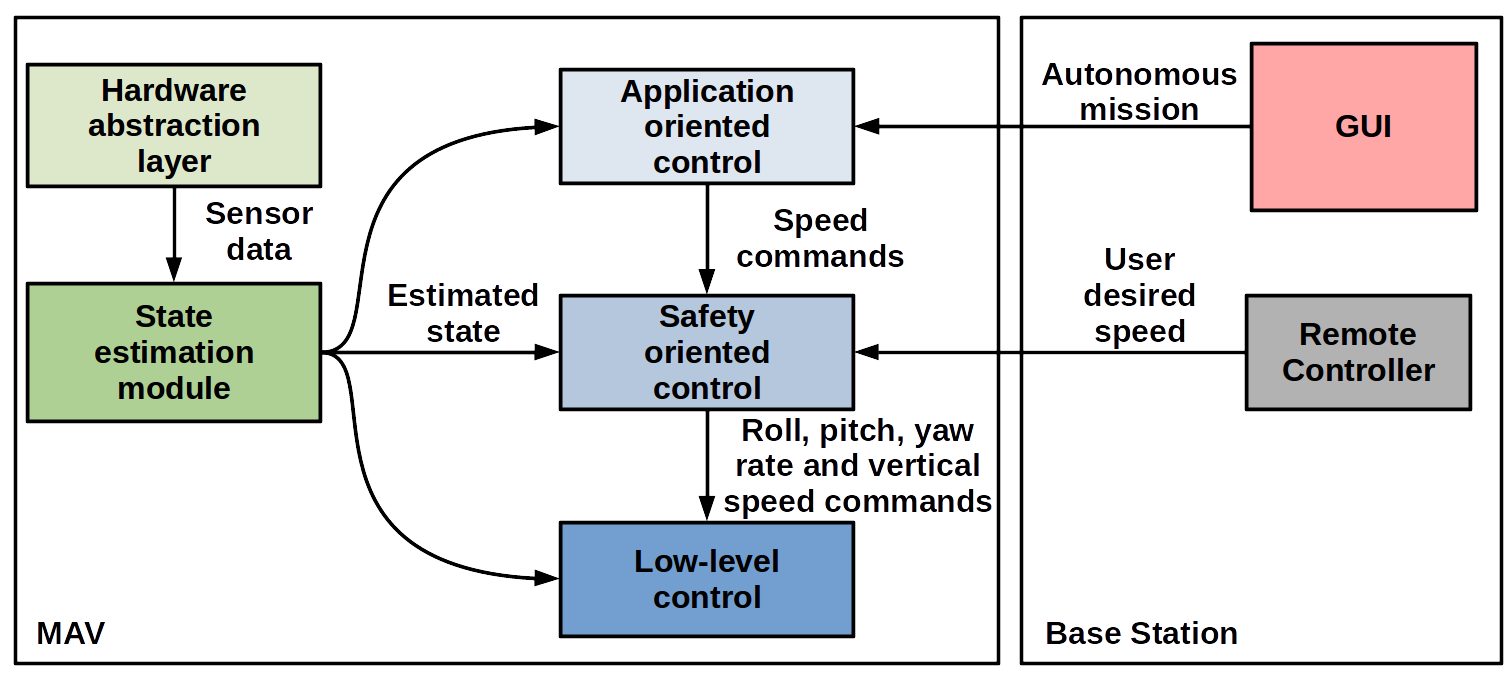}
\caption{Control architecture of the MAV.}
\label{fig:controlarch}
\end{figure}


The control architecture \hide{of MUSSOL} has been designed following open-source principles and using the Robot Operating System (ROS) as middleware. The control software is structured in different layers, organized hierarchically, as shown in Fig.~\ref{fig:controlarch}. Each of these layers implements a different control level that contributes to generate the final command to be issued to the platform. Further information about the different layers can be found next.

\subsection{State Estimation}

The state estimation module, transversal to all layers, is in charge, as its name suggests, of calculating as accurate as possible the MAV state by means of fusing all data received from the available sensors. In detail, the vehicle state comprises the platform pose $(x,y,z,\varphi,\theta,\psi)$, the linear velocities $(\dot{x},\dot{y},\dot{z})$ and accelerations $(\ddot{x},\ddot{y},\ddot{z})$, and the angular velocities $(\dot{\varphi},\dot{\theta},\dot{\psi})$. 

The estimation procedure starts by pre-processing sensor data (IMU, laser and height) to counteract biases and perform roll and pitch compensation, depending on the sensor. Next, the module estimates: (1) the platfom 2D roto-translation between two consecutive scans using a laser scan matcher; and (2) a filtered height and Z velocity of the platform, estimated by means of Kalman filtering. The resulting velocities in all axes, along with the IMU accelerations, are fed into a velocity estimation module to generate a final 3D speed estimate.


For inspection tasks, it is beneficial to be able to associate a 3D position to the data collected as well as to ensure a proper surface coverage about the surface under inspection. In order to implement these functionalities, the platform needs to estimate a full pose, i.e. not only its speed but also its 3D position. Due to this reason, this layer includes a module that fuses, in a flexible way, navigation data and position sources to calculate the absolute pose of the vehicle. This is done by combining two Extended Kalman Filters (EKF): a \textit{local} EKF, which fuses local position estimates, and a \textit{global} EKF, which combines absolute position estimates from sources such as GPS (only if available), Ultra-Wide Band (UWB) localization system and laser/vision-based Simultaneous Localization and Mapping (SLAM) solutions.

\subsection{Low-Level Control}

This level is in charge of controlling vertical motion and accomplishing attitude stabilization and direct motor control. Functionalities related to this level are executed entirely in the FMU. The software architecture sends motion commands, defined in terms of roll, pitch, vertical velocity and yaw rate setpoints, from within the corresponding control loops (using the available functions from the Software Development Kit (SDK) provided by the manufacturer). Additionally, the SDK provides information about the platform, e.g. battery voltage, IMU data, and height from the barometric pressure sensor, all used at different points of the control architecture.

\subsection{Behaviour-based Architecture}

\begin{figure}[tb]
\centering
\includegraphics[width=1.0\linewidth]{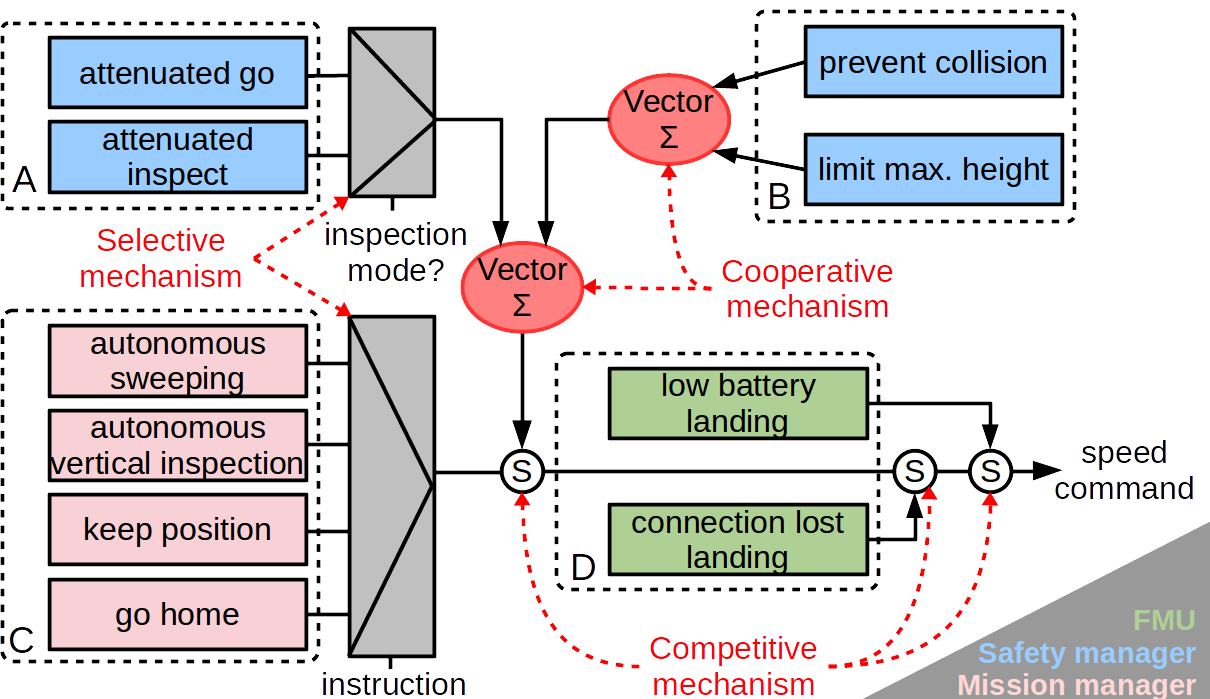}
\caption{Behaviour-based architecture: (A) behaviours to accomplish the user intention, (B) behaviours that ensure the platform safety within the environment, (C) behaviours that increase the autonomy level, and (D) behaviours oriented to check flight viability. The different combination mechanisms are indicated using red arrows.}
\label{fig:behavarch}
\end{figure}


Commands sent to the low-level control layer are generated by a set of robot behaviours organized in a hybrid competitive-cooperative framework~\cite{arkin1998}. In our control architecture, behaviours have been organized into four categories, according to its purpose for the visual inspection application: (A) behaviours to accomplish the user intention, (B) behaviours to ensure the platform safety within the environment, (C) behaviours to increase the autonomy level, and (D) behaviours to check flight viability. Figure~\ref{fig:behavarch} shows the behaviour-based architecture and how each behaviour contributes to the final command. Behaviours in the A, B and C categories run either within the \textit{Safety-oriented control} and the \textit{Application-oriented control} layers. The last category of behaviours (D) are in charge of ensuring that the flight can start or progress during the mission by checking different situations that can compromise the platform safety, such as, for instance, losing connectivity with the base station or detecting low battery. These behaviours are mainly executed within the FMU. These modules are detailed in the next sections.

\subsection{Safety-Oriented Control}


This layer introduces mechanisms to ensure the integrity of the MAV, generating safe commands to manage the platform during an inspection mission. It comprises two modules, namely the \textit{Safety manager} and the \textit{Flight controller}.

The \textit{Safety manager} module combines the output of the behaviours corresponding to categories A and B (see Fig.~\ref{fig:behavarch}), in order to accomplish the user intention while, at the same time, preserving the platform safety. These behaviours are defined next:

\begin{itemize}


    \item \textit{Behaviours to accomplish the user intention}. This category consists of the \textit{attenuated go} and \textit{attenuated inspect} behaviours. On the one hand, the \textit{attenuated go} behaviour conveys the user desired speed vector but attenuated according to the proximity of the platform to surrounding obstacles. On the other hand, the \textit{attenuated inspect} behaviour proceeds in the same way but only when the \textit{inspection mode} is activated. This mode limits the maximum speed of the vehicle and the distance to the inspected surface, which keeps the platform at a close distance to the front surface under inspection, enhancing the quality of the gathered data.
 

     \item \textit{Behaviours to ensure the platform safety within the environment}. This category includes the \textit{prevent collision} behaviour, which generates a repulsion vector whose magnitude is a function of the distance for each obstacle near the MAV, and a \textit{limit max height} behaviour to avoid the platform from flying above a predefined maximum height.
   
\end{itemize}


The \textit{Flight controller} module generates the final command to be issued to the platform. It is implemented as a finite state machine consisting of four states: \textit{on ground}, \textit{taking off}, \textit{flying} and \textit{landing}. During the \textit{flying} stage, two PID controllers keeps the speed command in longitudinal and lateral axes, while the vertical motion control runs inside the FMU. When the vertical speed command is zero, an additional PID runs to keep the platform at a constant height.

\subsection{Application-Oriented Control}


The main goal of this layer is to increase the autonomy level of the platform by means of inspection-oriented autonomous missions. Missions are described in a qualitative way, e.g. sweep the front wall from end to end, using the Graphical User Interface (GUI), in accordance to the SA paradigm. Internally, this layer transforms user-specified missions into a sequence of platform motion commands. For now, autonomous inspection missions are defined in terms of waypoints, although it is intended to be highly extensible, to be able to incorporate additional autonomy, in any form, as it is available. The \textit{Mission manager} and the \textit{Position controller} modules make up this layer.


The \textit{Mission manager} module is defined in terms of behaviours belonging to category C (see Fig.~\ref{fig:behavarch}). It consists of four behaviours named \textit{autonomous sweeping}, \textit{autonomous vertical inspection}, \textit{keep position} and \textit{go home}.


The \textit{autonomous sweeping} behaviour generates a sequence of waypoints that allows the platform to sweep a rectangular area parallel to the surface under inspection. The dimensions of the sweeping can be either specified by the user through the GUI, or automatically computed by the vehicle when the user enables an end-to-end sweeping of an entire wall. In the same way, the \textit{autonomous vertical inspection} behaviour issues a set of waypoints to inspect a vertical structure from both sides. The vehicle ascends inspecting the structure from the left and then descends inspecting the vertical structure form the right. The \textit{keep position} behaviour keeps the vehicle at the current position. Finally, the \textit{go home} behaviour makes the platform return to the home position. This position is defined automatically after taking-off, although the user can redefine it through the user interface.


The sequence of waypoints generated by the \textit{Mission controller} are sent to the \textit{Position controller} module, which generates the suitable speed commands to attain the different waypoints related to the mission accomplishment. This is done by means of PID-based position control. The resulting commands are consolidated within the \textit{Safety manager} module, taking into account other sensor data, to preserve the platform integrity.





\section{Experimental Results}
\label{sec:expres}


This section reports on an extensive set of experimental results that show the performance level of both the control architecture and the underlying vehicle. All the experiments were performed inside indoor or semi-indoor environments and over the real platform, or within a Hardware-in-the-Loop (HiL) simulation framework using platform sensors for state estimation. This section comprises two main parts. The first part focus on the evaluation of the low-level capabilities of the system, namely, hovering, collision avoidance and the capability to reach a given waypoint. These experiments were performed in our laboratory, where a motion tracking system is available and can be used to get high-accuracy ground truth data. In the second part, we focus on evaluating the higher-level capabilities of the platform, assessing the autonomous sweeping and autonomous vertical inspection functionalities, first in simulation, and next on-board real vessels. Each set of experiments is detailed in the following sections.

\subsection{Low-Level Skills Evaluation}

\begin{figure}[tb]
\centering
\begin{tabular}{@{\hspace{0mm}}c@{\hspace{0mm}}c@{\hspace{0mm}}}
\includegraphics[width=0.60\linewidth,clip=true,trim=9mm 12mm 10mm 0]{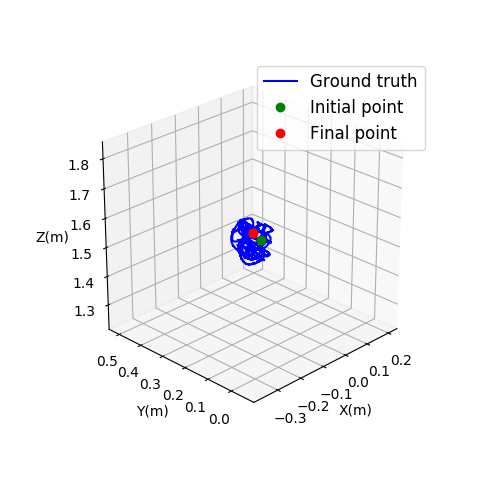} & 
\includegraphics[width=0.40\linewidth,clip=true,trim=4mm -7mm 8mm 0]{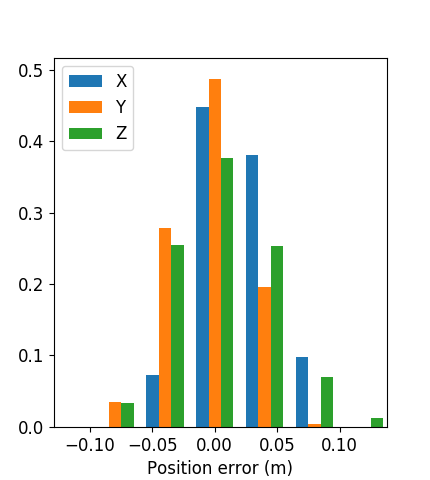} \\
\end{tabular}
\caption{Results for a 1-minute hovering: (left) positions provided by the motion tracking system, (right) estimated probability density functions of the position error for each axis.}
\label{fig:exp_lab_hovering}
\end{figure}

As a first experiment, Fig.~\ref{fig:exp_lab_hovering} shows a one-minute hovering, which starts after issuing a keep position command to the platform. The left plot shows the position of the MAV provided by the motion tracking system, while the right figure illustrates the error against the commanded position for each axis using a histogram representation. All three histograms are estimations of the Probability Density Function (PDF) of the error for each axis. As it is shown, the PDFs are all centered at zero and most part of the probability keeps confined within the interval $\pm$5 cm with respect to the issued waypoint.

\begin{figure}[tb]
\centering
\includegraphics[width=1.0\linewidth,clip=true,trim=8mm 0 15mm 0]{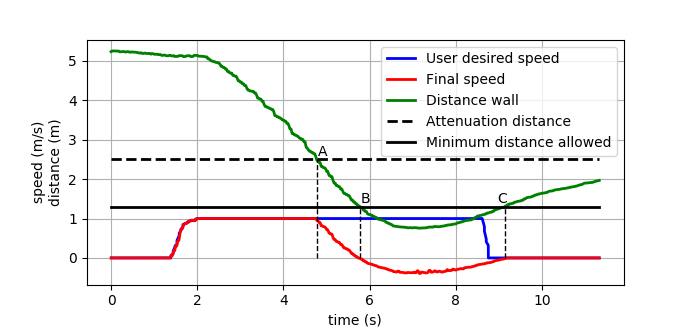}
\caption{Performance of the behaviours implementing the collision avoidance capability.}
\label{fig:exp_lab_behav}
\end{figure}


In a second experiment, we assess the operation of the behaviours related to obstacle avoidance, which are part of the \textit{Safety manager} module. In this regard, results for the joint action of the \textit{attenuated go} and the \textit{prevent collision} behaviours can be found in Fig.~\ref{fig:exp_lab_behav}. First, a command to move the platform against a wall is issued to the MAV. As shown in the figure, the longitudinal speed command produced by the \textit{Safety manager} module coincides with the user command until the distance to the wall is lower than 2.5 m (instant A). At this point, the user-desired velocity is reduced by the \textit{attenuated go} behaviour decreasing the speed according to the current distance to the wall. When the wall is closer than 1.3 m (instant B), which has been set as the minimum allowed distance to obstacles, the user command is aborted by the \textit{prevent collision} behaviour and the final speed command issued to the MAV becomes negative to separate it from the wall (instant C).

\begin{figure}[tb]
\centering
\begin{tabular}{@{\hspace{0mm}}c@{\hspace{0mm}}c@{\hspace{0mm}}}
\includegraphics[width=0.55\linewidth, clip=true, trim=50 50 50 50]{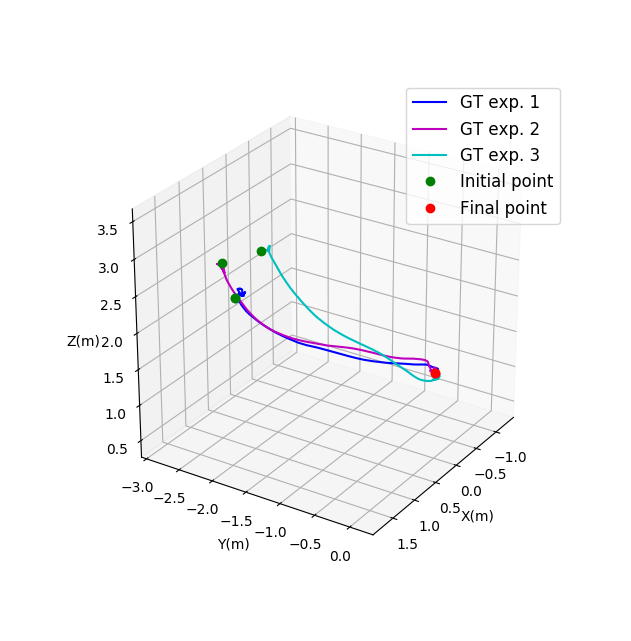} & 
\includegraphics[width=0.45\linewidth, clip=true, trim=10 -100 30 0]{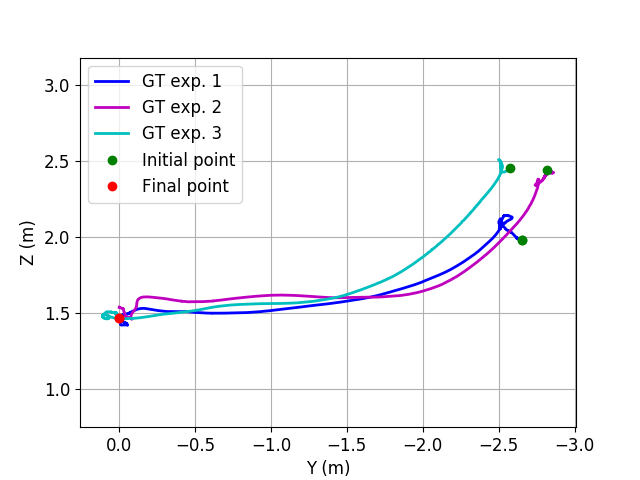} \\
\end{tabular}
\caption{Paths followed by the MAV when testing the capability to attain a waypoint. Results correspond to three different executions trying to reach waypoint [0.0, 0.0, 1.5].}
\label{fig:exp_lab_go_home}
\end{figure}

As part of the evaluation of the platform low-level skills (under laboratory conditions), we next report results for a third experiment to asses the platform performance while trying to reach a given waypoint. This capability is crucial to accomplish waypoint navigation as accurate as needed by the \textit{application-oriented behaviours}, such as the \textit{autonomous sweeping}. Figure~\ref{fig:exp_lab_go_home} shows the paths followed by the MAV for three different executions of the \textit{go home} behaviour. As can be observed, the position controllers always manage to return the MAV to the home position situated at coordinates [0.0, 0.0, 1.5] m.

\subsection{High-Level Skills Evaluation}


As part of the validation of the high-level skills of the platform, we have performed experiments in two different environments. First, we have evaluated the system within a complex but simulated scenario, used to adjust and configure the architecture in a safe way. Next, we report on similar experiments during field trials on board a real vessel, and illustrate the system working on a real environment. Details about these experiments can be found in the following sections.

\subsubsection{Experiments within a Simulated Environment}


To show the performance of the MAV within a simulated environment, we make use of Gazebo, which is a well-known simulator fully integrated within the ROS framework, so that the full control architecture of the MAV can directly interact with the simulated environment. The high-level skills of the platform, namely \textit{autonomous sweeping} and \textit{autonomous vertical inspection}, are evaluated in simulation not only because the simulation environment allows to debug errors in the code, but also because the simulation environment allows performing autonomous sweepings (and autonomous vertical inspections) whose dimensions can be much larger than the inspection missions that can be run inside our laboratory. 

\begin{figure}[tb]
\centering
\begin{tabular}{@{\hspace{0mm}}c@{\hspace{0mm}}c@{\hspace{0mm}}}
\includegraphics[width=0.50\linewidth, clip=true, trim=0 0 0 0]{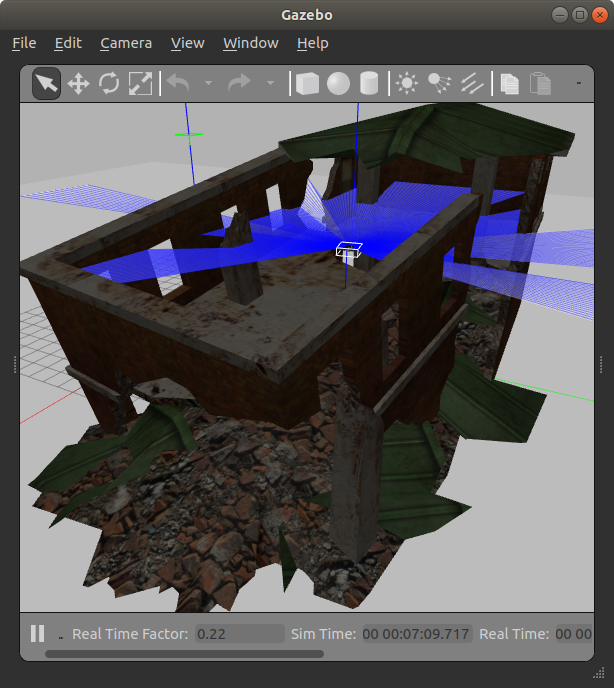} & 
\includegraphics[width=0.50\linewidth, clip=true, trim=0 0 0 0]{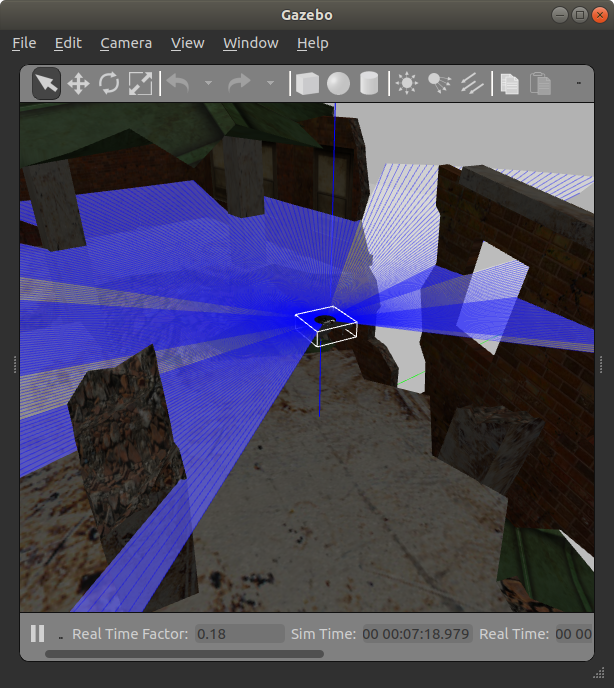} \\
\end{tabular}
\caption{Screenshots of the simulation environment used to test the high-level skills of the MAV.}
\label{fig:exp_sim_gazebo}
\end{figure}

To perform the experiments, we have loaded a freely available environment comprising a collapsed fire station, to be inspected by means of the simulated MAV. Figure~\ref{fig:exp_sim_gazebo} shows two screenshots of the Gazebo tool simulating the MAV flying inside the collapsed building. The blue lines coming out from the aerial vehicle simulate the ranges emitted by the laser scanner.

\begin{figure*}[tb]
\centering
\begin{tabular}{@{\hspace{0mm}}c@{\hspace{0mm}}c@{\hspace{0mm}}c@{\hspace{0mm}}c@{\hspace{0mm}}}
\includegraphics[width=0.50\columnwidth, clip=true, trim=50 50 75 50]{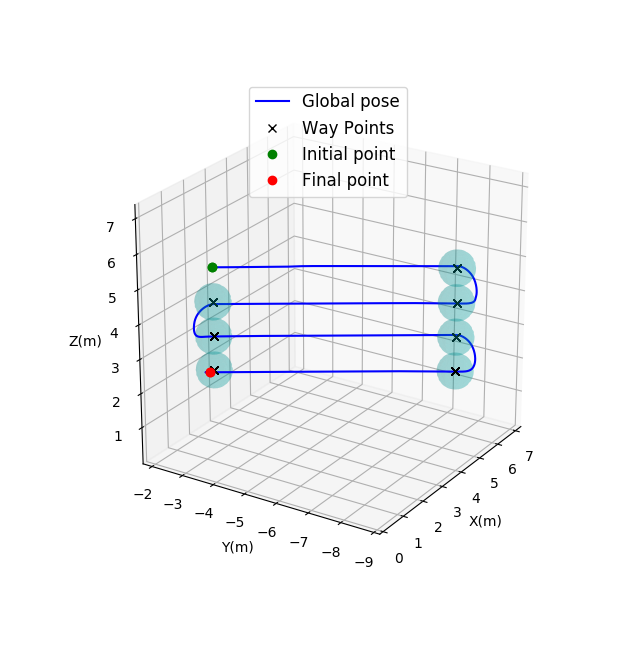} & 
\includegraphics[width=0.50\columnwidth, clip=true, trim=50 50 80 50]{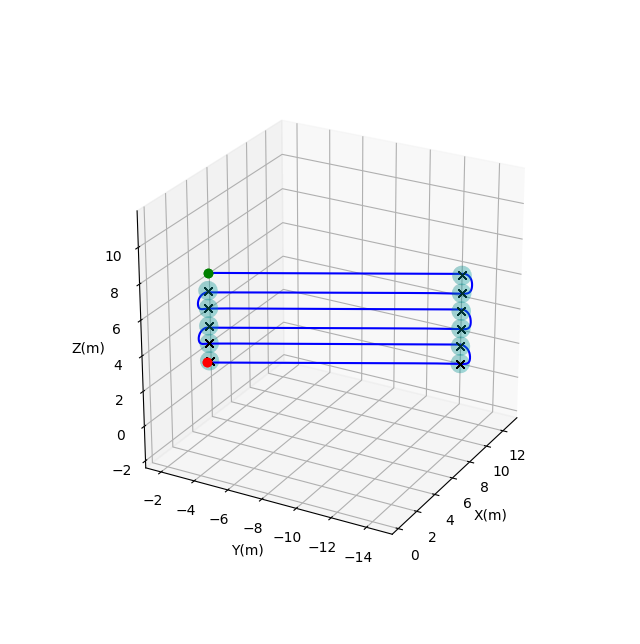} & 
\includegraphics[width=0.50\columnwidth, clip=true, trim=60 50 90 50]{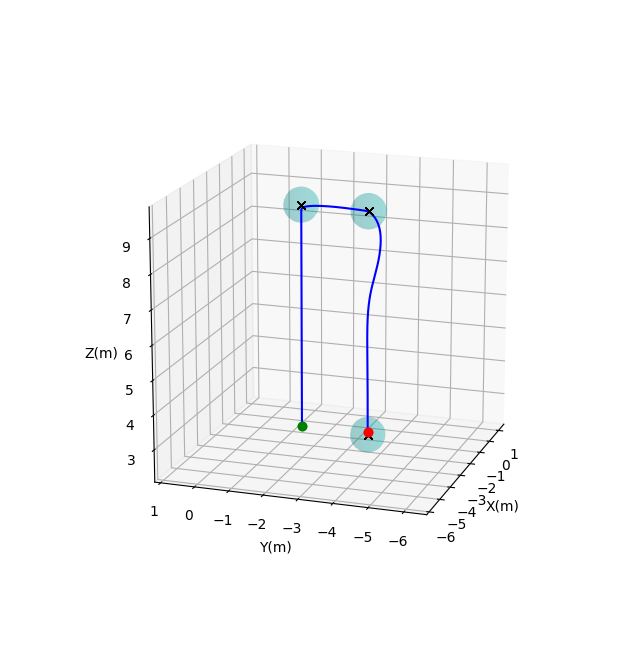} &
\includegraphics[width=0.50\columnwidth, clip=true, trim=60 50 80 50]{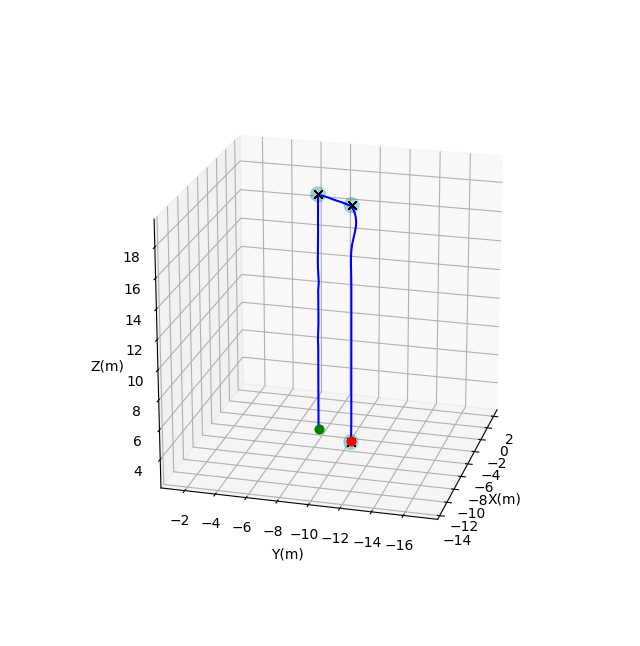} \\
\end{tabular}
\caption{Paths followed by the simulated MAV when performing two autonomous sweepings (left) and two autonomous vertical inspections (right).}
\label{fig:exp_sim_results}
\end{figure*}

Figure~\ref{fig:exp_sim_results} shows the different paths followed by the simulated MAV while performing the inspection of the collapsed fire station. The two left-most plots correspond to the vehicle while running two autonomous sweeping tasks of, respectively, 6 $\times$ 3 and 13 $\times$ 5 m, while the two right-most plots show the trajectories corresponding to, respectively, two vertical inspections at maximum heights of 6.5 and 16 m. The resulting waypoints generated by the mission manager are indicated using 50-cm light blue spheres. These spheres correspond to the tolerance volume used for the achievement of the waypoints.


\subsubsection{On-board Field Trials}

\begin{figure}[tb]
\centering
\begin{tabular}{@{\hspace{0mm}}c@{\hspace{1mm}}c@{\hspace{0mm}}}
\includegraphics[width=0.48\linewidth, clip=true, trim=500 0 500 0]{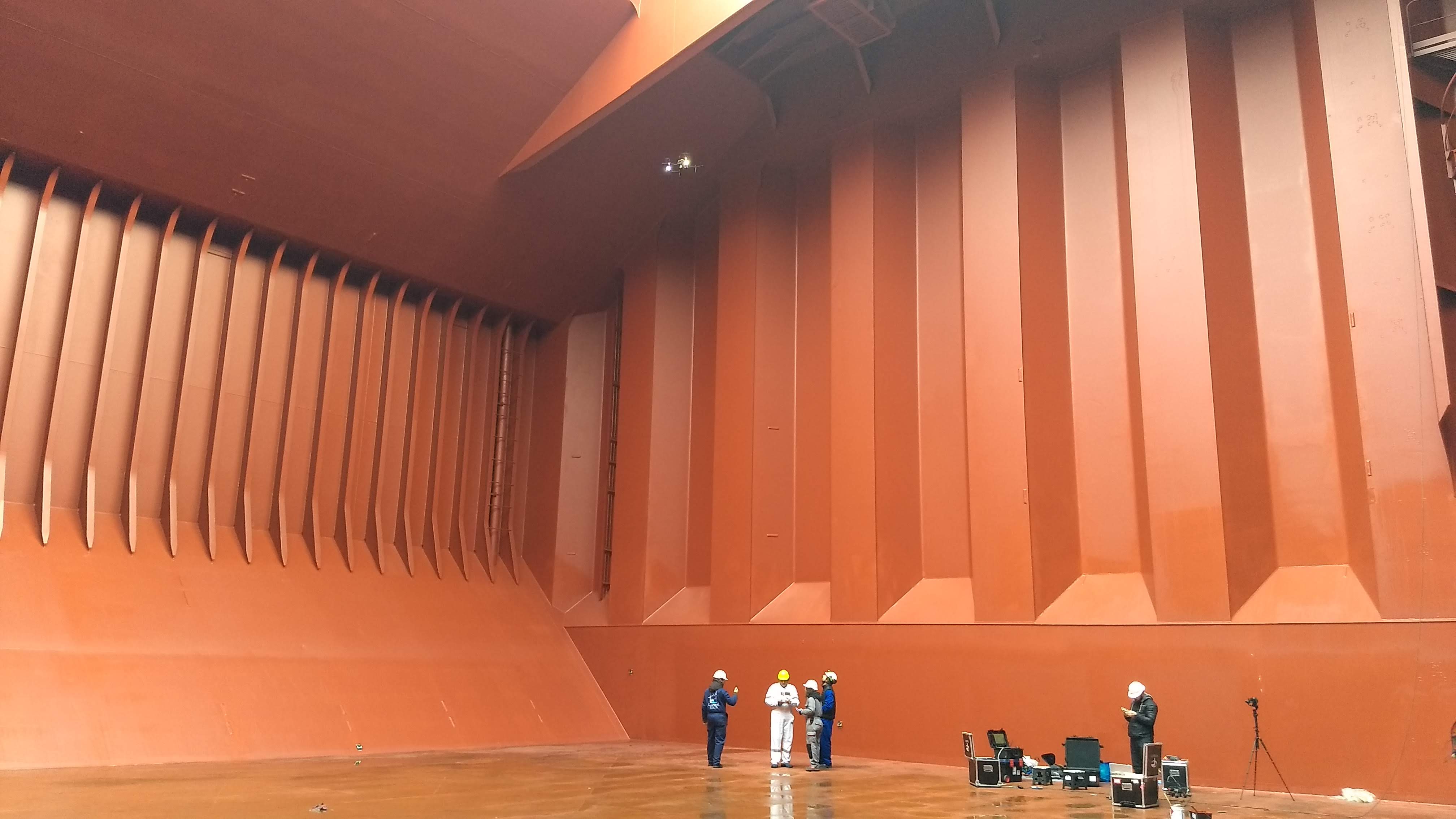} & 
\includegraphics[width=0.48\linewidth, clip=true, trim=500 0 500 0]{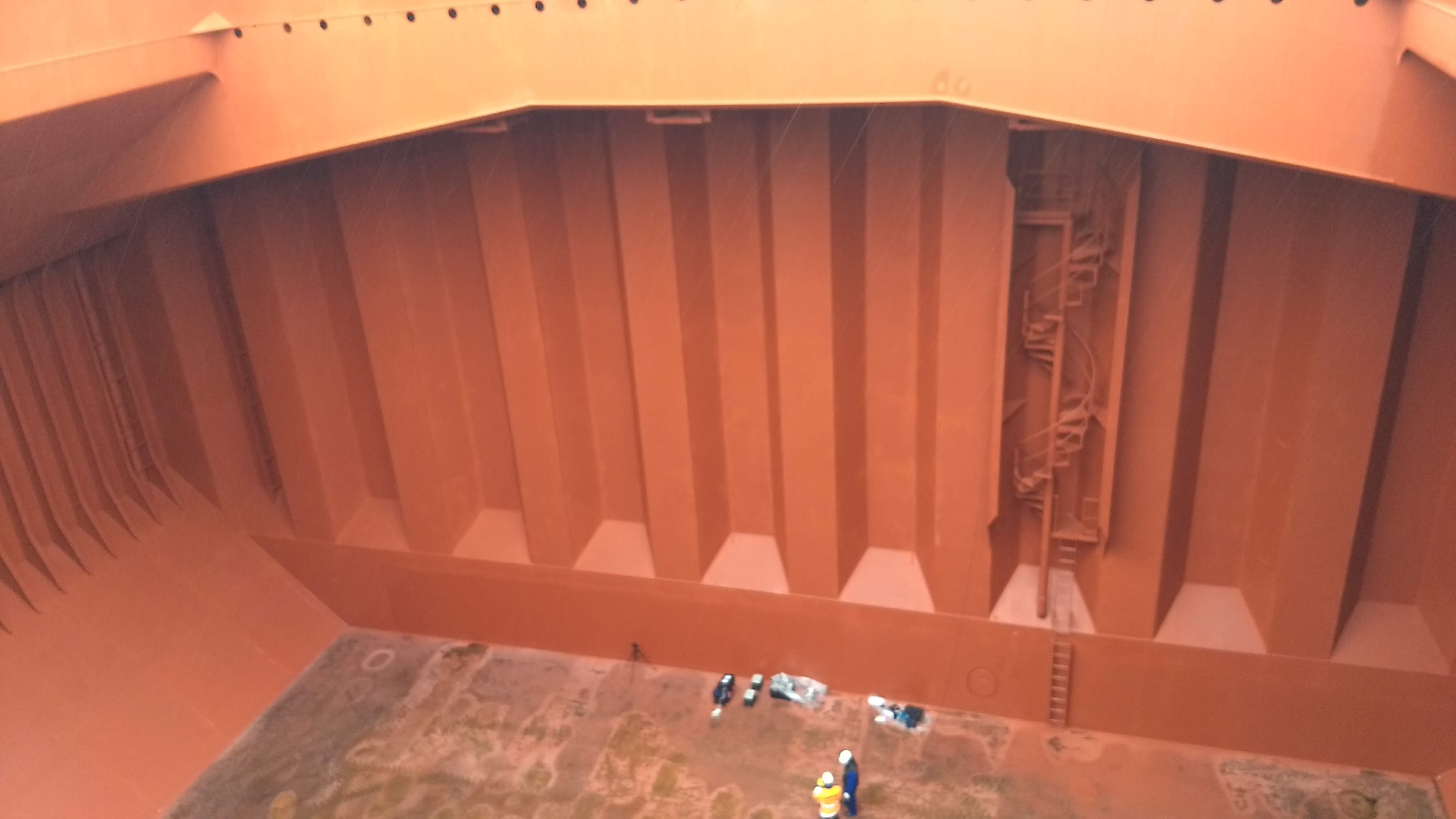} \\
\includegraphics[width=0.48\linewidth, clip=true, trim=500 0 500 0]{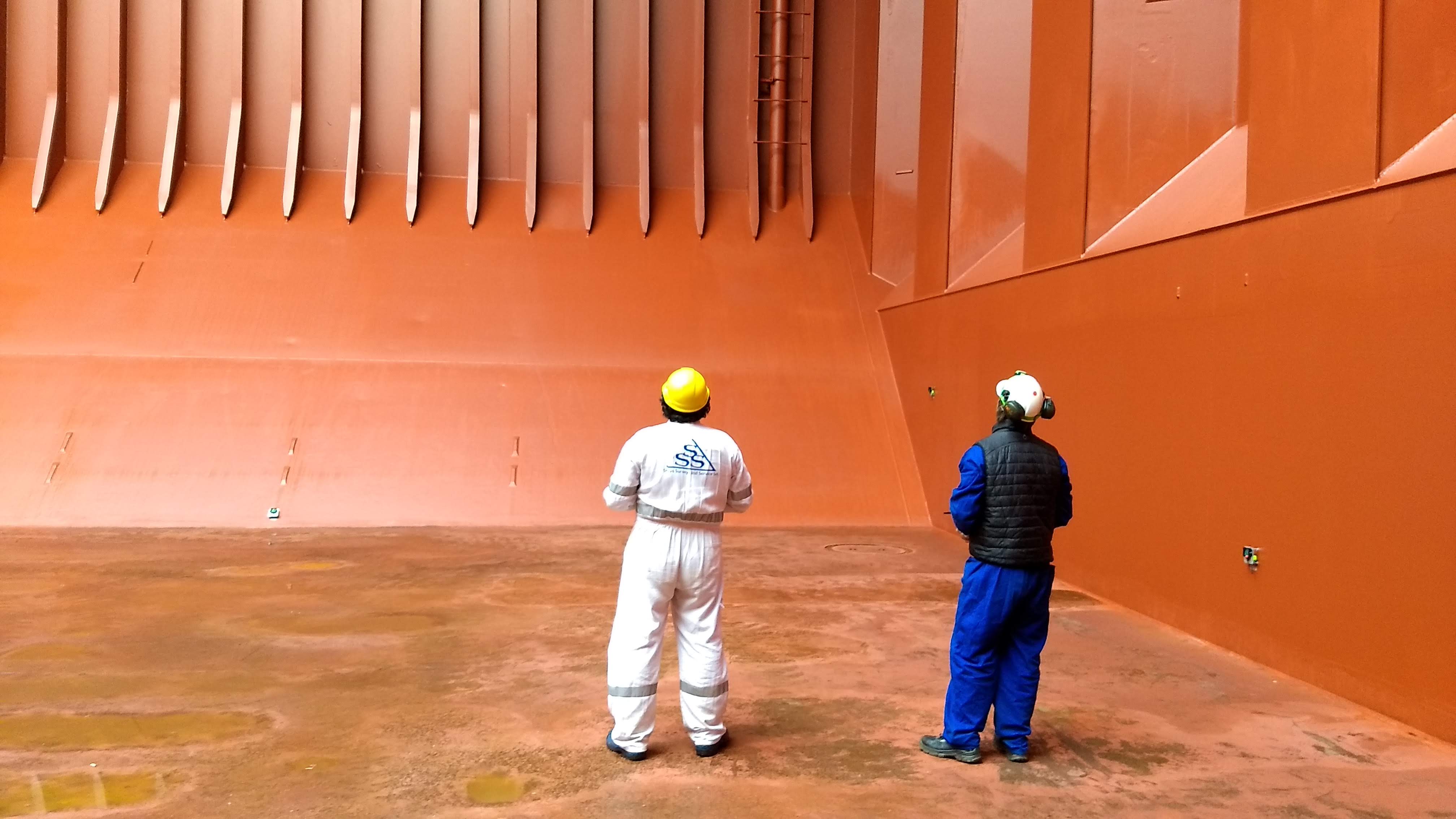} & 
\includegraphics[width=0.48\linewidth, clip=true, trim=0 1850 0 500]{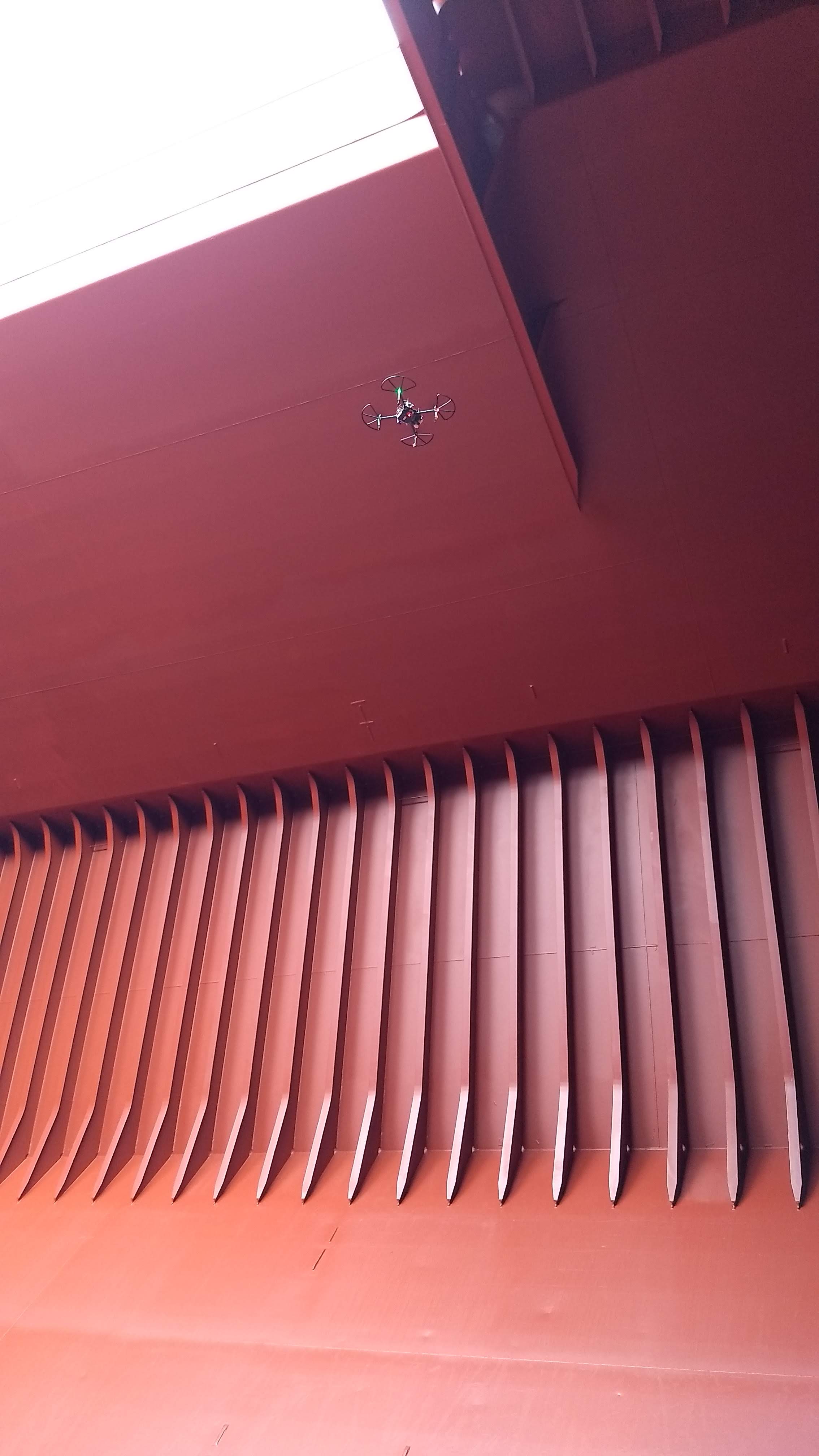} \\
\end{tabular}
\caption{Pictures of the field trials taking place on-board a bulk carrier.}
\label{fig:exp_bulkcarrier}
\end{figure}

\begin{figure*}[tb]
\centering
\begin{tabular}{@{\hspace{0mm}}c@{\hspace{0mm}}c@{\hspace{0mm}}c@{\hspace{0mm}}}
\includegraphics[width=0.3\linewidth, clip=true, trim=0 0 0 0]{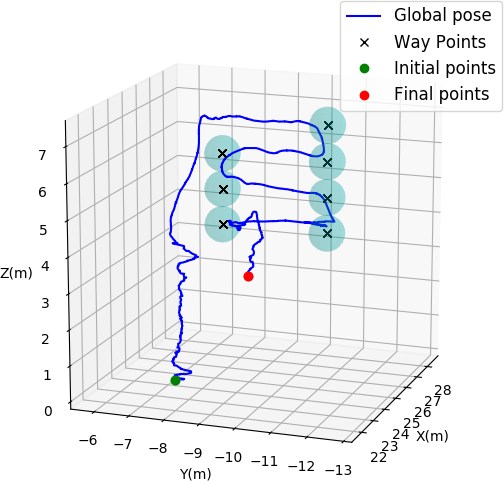} & 
\includegraphics[width=0.3\linewidth, clip=true, trim=0 0 0 0]{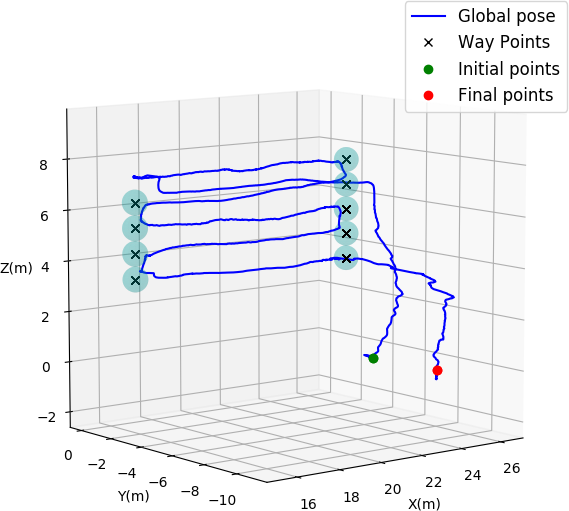} & 
\includegraphics[width=0.39\linewidth, clip=true, trim=0 0 0 90]{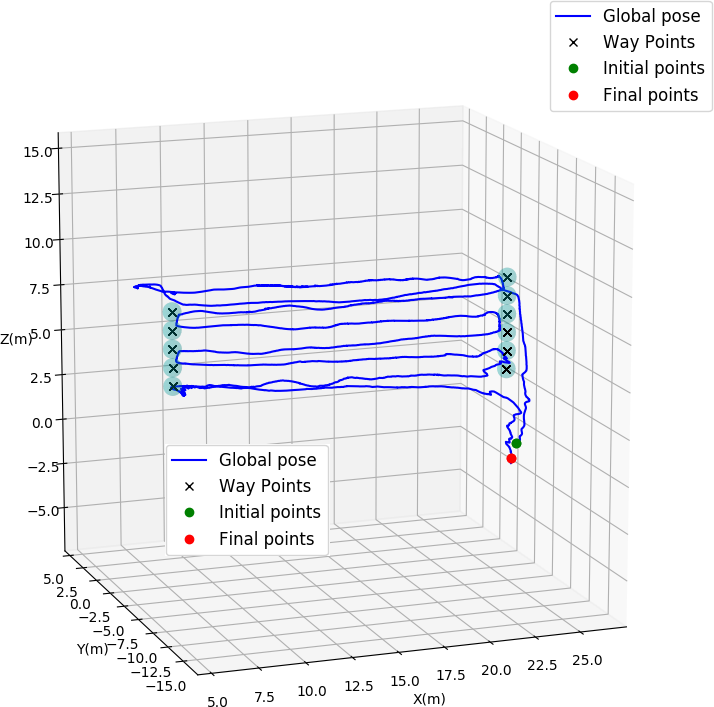} \\
\end{tabular}
\caption{Paths estimated for different autonomous sweepings performed on-board a bulk carrier.}
\label{fig:exp_vessels_res}
\end{figure*}


In order to validate the whole system in a real environment, we performed several experiments inside a cargo hold of a bulk carrier (see Fig.~\ref{fig:exp_bulkcarrier}). the platform was used to inspect several structures of the cargo hold according to the indications of an experienced surveyor, who, during the flight, managed a secondary R/C to operate the MAV camera. It is worth noting that the hatch of the cargo hold was open during the experiments, so that the MAV was affected by the effect of the wind, typically present in harbour areas.


Regarding the performance level of the platform while as for the evaluated high-level skills, Fig.~\ref{fig:exp_vessels_res} shows the paths followed by the vehicle during the above-mentioned experiments. The left and middle plots correspond to sweepings whose dimensions have been previously specified by the user through the GUI, while the rightmost plot corresponds to an end-to-end autonomous sweeping flight covering all the frames of the port side wall of the cargo hold (approx. 18 m). As done before for the simulation experiments, waypoints are indicated using 50-cm light blue spheres. Unlike the plots for the simulation environments, these plots show the entire paths followed by the MAV, from take-off until landing, i.e. not only the autonomous sweeping. Notice that these plots correspond to the position estimated by the state estimation module. As can be observed, the MAV is able to carry out the sweeping task despite the external disturbances from the wind. 

\section{Conclusions and Future Work}
\label{sec:conclusions}



This paper describes and reports on the evaluation results of a control architecture for aerial vehicles focusing on visual inspection applications. This control architecture, specifically designed and developed around the Supervised Autonomy (SA) paradigm, integrates a number of suitable high-level control behaviours aiming at simplifying and making visual inspections more cost-effective and safer. The large-tonnage vessel inspection problem has been taken as the use case for this work. 

The control architecture runs on a multi-rotor platform properly fitted, in terms of equipment, to satisfy the scope of the survey activities within the intended scenario. Following the SA paradigm, the MAV is complemented with a Base Station (BS), comprising in turn a Ground Control Unit (GCU) with \textit{in-situ} data processing capabilities, and suitable Hardware Interaction Devices (HID) to facilitate the interaction with the MAV and collect/process inspection data during flight.


The skills of the platform have been evaluated through a complete set of experiments, including laboratory tests, simulation evaluations and field trials on board a real vessel. The experimental evaluation has covered the different levels of complexity related to the operation of the platform, starting with the assessment of the lower-level skills and finishing with the higher-level capabilities, focused on increasing the autonomy of the robotic platform.


Future work includes the incorporation of alternative sensors such as e.g. a lightweight 3D laser scanner to enhance the perception of the environment and attain new functionalities within the inspection scope. In the same line, we plan to enhance and extend the control architecture by developing and incorporating additional high-level capabilities that can increase the usefulness of the platform as for visual inspection, e.g. explore unknown, confined spaces and/or survey specific structures.



\bibliographystyle{IEEEtran}
\bibliography{ms}

\end{document}